\documentclass[10pt,twocolumn,letterpaper]{article}

\usepackage{cvpr}              

\usepackage{graphicx}
\usepackage{amsmath}
\usepackage{amssymb}
\usepackage{booktabs}

\usepackage[pagebackref,breaklinks,colorlinks]{hyperref}

\usepackage[capitalize]{cleveref}
\crefname{section}{Sec.}{Secs.}
\Crefname{section}{Section}{Sections}
\Crefname{table}{Table}{Tables}
\crefname{table}{Tab.}{Tabs.}

\def\cvprPaperID{6410} 
\def\confName{CVPR}
\def\confYear{2022}

\begin{document}

\title{FooDI-ML: a large multi-language dataset of food, drinks and groceries images and descriptions}

\author{David Amat Olóndriz\\
{\tt\small david.amat@glovoapp.com}
\and
Ponç Palau Puigdevall\\
{\tt\small ponc.puigdevall@glovoapp.com }
\and
Adri\`a Salvador Palau\\
glovoapp\\
Carrer de Pallars, 190, 08005 Barcelona\\
{\tt\small adria.salvador@glovoapp.com  }
}

\maketitle

\begin{abstract}
In this paper we introduce the FooDI-ML dataset. This dataset contains over 1.5M unique images and over 9.5M store names, product names descriptions, and collection sections gathered from the Glovo application. The data made available corresponds to food, drinks and groceries products from 37 countries in Europe, the Middle East, Africa and Latin America. The dataset comprehends 33 languages, including 870K samples of languages of countries from Eastern Europe and Western Asia such as Ukrainian and Kazakh, which have been so far underrepresented in publicly available visio-linguistic datasets. The dataset also includes widely spoken languages such as Spanish and English. To assist further research, we include benchmarks over two tasks: text-image retrieval and conditional image generation.
\end{abstract}

\section{Introduction}
\label{sec:intro}

The COVID19 pandemic has accelerated the digitalisation of restaurants and the growth of the food delivery sector. National lockdowns have made it impossible to go to bars and restaurants, which has prompted many people to discover the possibility of ordering food and drinks online. Therefore, solving tasks such as text-image retrieval for food and drinks search engines has become increasingly important. The lack of large-scale multilingual datasets covering this domain \cite{marin2019recipe1m+} means that it is hard to build efficient search engines and recommender systems, especially for under-represented languages \cite{ramkishoresaravanan_2021}. In this paper, we aim to help bridge this gap by offering a large multi-language dataset that covers several of such languages, in addition to some common languages such as Spanish, French and English: the FooDI-ML dataset. Our dataset features data collected by a food and groceries delivery app: \textbf{Glovo}.

In recent years there has been increased attention to visio-linguistic datasets. Besides classical vision tasks such as image classification and segmentation, image and text datasets can be used to learn multi-modal tasks. For example, image captioning, text to image generation and text-image retrieval.  The size of publicly available image and text datasets has grown steadily from the flickr30K dataset \cite{young2014image}, including 30K English-language examples, to the WIT dataset, a multi-language dataset extracted from Wikipedia that includes over 11M unique examples. To our knowledge, FooDI-ML is the second largest multilingual visio-linguistic dataset available to date. 

The structure of this paper is as follows: in \cref{sec:prior_work} we review prior work and place our dataset in a broader research context. In \cref{sec:the_dataset} we describe the dataset in detail, including notable examples and proposed tasks. \cref{sec:benchmark_task} presents benchmarks over the text-image retrieval and conditional image generation tasks.  Finally, \cref{sec:conclusion} features the conclusion and a discussion on future work. 


\subsection{Our contribution}
\begin{itemize}
    \item We present the FooDI-ML dataset, a dataset of over 2.8M food, drinks and grocery images included together with over 9.5M natural language descriptions, store names, product names, and collection sections in which they are included. It is the second largest multilingual dataset of this kind publicly available.
    \item We provide a dataset covering a use domain rarely covered (multi-language samples for food, drinks and groceries).
    \item The dataset covers languages that despite having a large number of speakers are under-represented in trained datasets. Therefore, it has potential to reduce bias in food, drinks and grocery based search engines.
    \item We compute a train/test/validation split and two benchmark tasks to facilitate future research.
\end{itemize}
\section{Prior work}
\label{sec:prior_work}
\subsection{Datasets}
Datasets including images coupled to natural language data are often referred to as visio-linguistic datasets \cite{srinivasan2021wit}. The two pioneering datasets in the field were published just a few months apart from each other: the flickr30K and COCO datasets \cite{young2014image,lin2014microsoft}. Both datasets were collected by asking humans to label tenths of thousands of images manually. This means that these datasets, albeit seminal, do not include an amount of data comparable with vision-only or text-only datasets with sample numbers routinely reaching the order of 100M samples or more \cite{thomee2016yfcc100m,lin2014microsoft,sun2017revisiting,yalniz2019billion}.

In addition to their relatively smaller size, these initial datasets included exclusively English language examples. This shortcoming spun an effort to extend the work to other languages. Soon, new datasets, together with relabeled versions of the same datasets appeared in several languages. Amongst them German \cite{hitschler2016multimodal}, Japanese\cite{P16-1168}, Chinese \cite{li2019coco}, Dutch \cite{van2018didec}, Italian \cite{antonio2019large} and Vietnamese \cite{lam2020uit}. In addition to this, crow-sourced translations of the COCO and flickr30K dataset are relatively frequent and can be found in non-peer-reviewed github repositories and other sources (see, for example, the Spanish translation available in Kaggle \cite{garcia_2020}).

The problem of dataset size and language coverage was not solved until recently when two approaches (and corresponding datasets) were introduced, both of them based in web crawling: the Conceptual Captions dataset \cite{sharma2018conceptual} and the Wikipedia-based Image Text Dataset \cite{srinivasan2021wit}. Google's Conceptual Captions (CC) dataset includes 3.3M English examples obtained by web crawling of images and alt-text annotations. The samples in the dataset are then cleaned using a complex funnel of rules \cite{sharma2018conceptual}. Another dataset from google, the google Open Images Dataset \cite{kuznetsova2020open} provides 9M images annotated with 675k localized narratives. The Wikipedia-based Image Text Dataset (WIT) managed to avoid complex cleaning rules by relying on Wikipedia as a highly curated data source. In doing so, it managed to gather a larger amount of high-quality data. This provided the largest publicly available dataset to date, including more than 11M unique examples in more than 100 different languages \cite{srinivasan2021wit}.

The increase in size and language coverage achieved by WIT and CC was especially aimed at obtaining large pretrained cross-modal networks. This, however, did not solve the problem of specificity. The lack of domain-specific datasets made it hard to train high-performant cross-modal networks for some applications. In our dataset's domain - food and drink images and descriptions - this was identified as a problem for the task of image recipe retrieval \cite{salvador2017learning}. In fact, the largest dataset available until recently contained only 101K samples pertaining to Chinese food recipes \cite{chen2016deep}. Other food datasets, such as the ISIA Food-500 dataset, contain up to 400K samples but their visio-linguistic value is limited, as each image is related to a class and a food name rather than to a natural language description \cite{min2020isia}. The FoodX-251 dataset (N=118K) \cite{kaur2019foodx}, ChineseFoodNet (180K) \cite{chen2017chinesefoodnet}, Food101 (101K) \cite{bossard2014food} and MAFood-121 (21K) \cite{aguilar2019regularized} are other notable examples of datasets targeting food classification. Only ChineseFoodNet is multilingual including both Chinese and English classes.

To fill in this void, Recipe1M+, a large dataset containing over 14M images and 1M recipes was made public for English-language recipes \cite{marin2019recipe1m+}. Despite the number of text samples being relatively small, the large number of food images was an important step towards improving machine learning tasks in the food domain. This dataset covers only one language and is focused in food recipes instead of food descriptions and broader food categories. The latter are more likely to appear in marketplace applications, and therefore might be more useful in an industrial context. 

To compare our dataset with other existing datasets we follow the approach of the WIT paper, where the number of languages and number of samples are the two main factors compared (see \cref{tab:table1}). Our dataset is the second biggest available both in terms of language and number of captioned samples.

\begin{table}
  \vspace{1mm}
  \centering
    \resizebox{\columnwidth}{!}{%
  \begin{tabular}{lllll}
    \toprule
    \cmidrule(r){1-2}
{Dataset} & {Images} & {Text} & {Langs.} & {Domain}  \\
    \midrule
    Flickr30K        & 32K             & 158K          & \textless{}8       & General      \\
    SBU Captions     & $\sim$1M        & $\sim$1M      & 1                  & General       \\
    MS-COCO          & $\sim$330K      & $\sim$1.5M    & \textless{}4       & General   \\
        CC               & $\sim$3.3M      & $\sim$3.3M    & 1                  & General  \\
    WIT              & 11.5M           & 37.6M         & 108                & Wikipedia   \\
        Open Images              & 9M           & 675K         & 1                & General   \\
        Recipe1M+              & 14M           & 1M         & 1                & Food   \\
        \textbf{FooDI-ML}         & \textbf{2.8M}            & \textbf{9.5M}           & \textbf{33}                 & \begin{tabular}[x]{@{}l@{}}\textbf{Food, drink}\\\textbf{groceries}\end{tabular}  \\
    \bottomrule
  \end{tabular}}
    \caption{Table showing the different visio-linguistic datasets. We choose to follow a similar format as \cite{srinivasan2021wit} for consistency. Note that our dataset is the second richest dataset in terms of text samples and languages. In this table, we quote the number of non-unique images.}
  \label{tab:table1}
\end{table}

\section{The FooDI-ML dataset}
\label{sec:the_dataset}
FooDI-ML is a dataset of food, drinks and groceries images and descriptions collected from all partners operating in the Glovo app in the last six years. The dataset contains data from 37 countries, with a significant representation of 33 languages. Amongst them are some common languages such as French, English and Spanish (which is the most common language), but also some rare languages such as Kazakh and Basque. More interestingly perhaps, some largely spoken languages such as Ukrainian but generally underrepresented in existing datasets are also present.

The dataset's size places it as the second richest publicly available visio-linguistic dataset, featuring  2.8M images (amongst them 1.5M unique), and up to 9.5M unique text samples. Each sample in the dataset corresponds to a different product that has been offered at some point in the Glovo app. There are 2.8M unique products (see \cite{app_2021} for detailed statistics). Each product has up to five data points associated with it: the store name, the product name, the product description, the product image and the product ``collection section". The latter is a meta category included in the app that restaurants and other sellers in the marketplace can use to organise their menu. Examples of collection sections are ``drinks", ``our pizzas", ``desserts", etc. Collection sections are not standardised, and can be in general chosen by the partner. Note that the number of products (2.8M) is larger than the number of unique images (1.5M), as repeated images are only counted once (although they can be repeated many times over the data set).

\begin{figure}[ht]
  \centering
    \includegraphics[height=7cm]{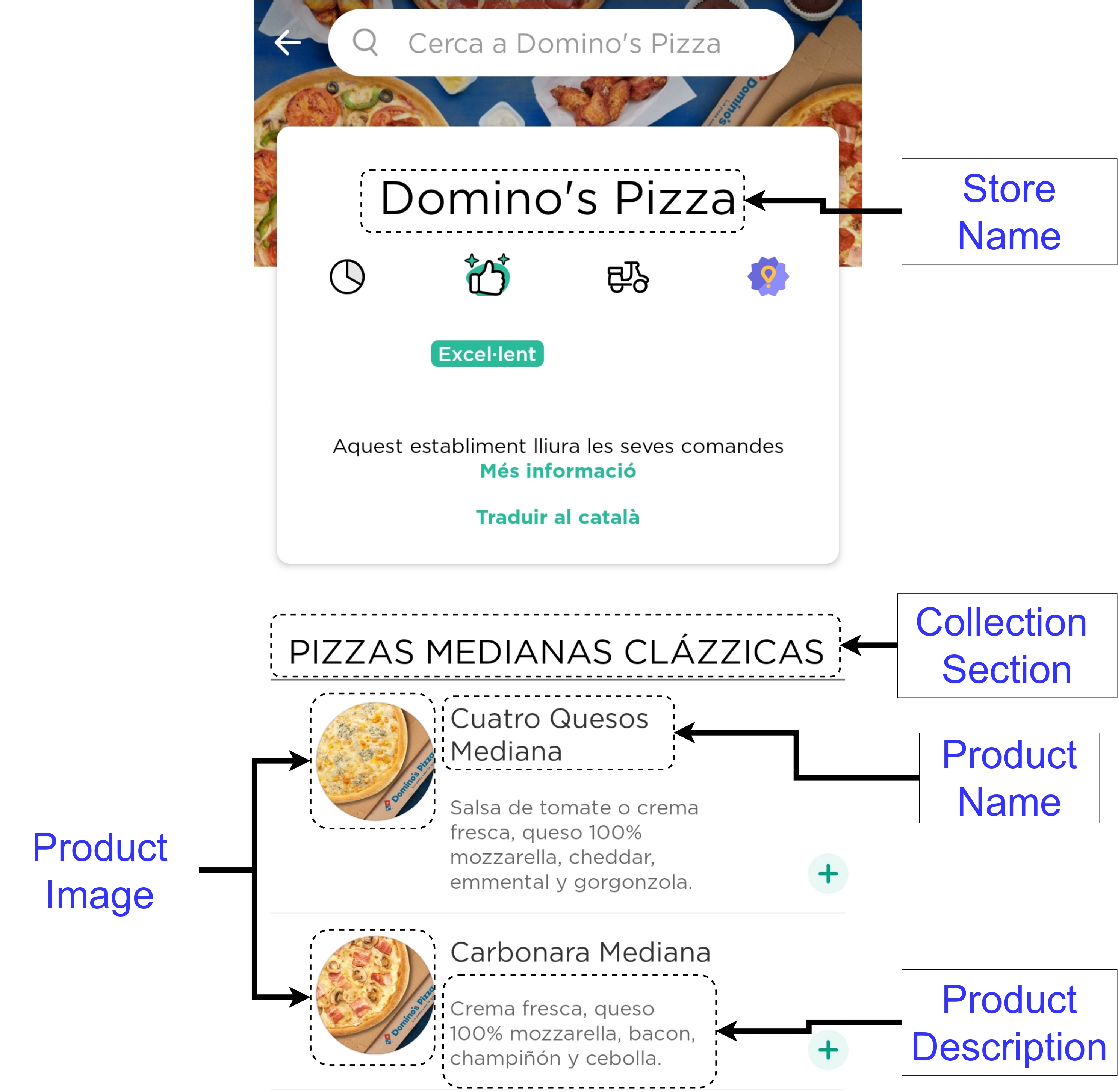}
  \caption{One of the samples included in our dataset as shown in the Glovo app (prices omitted). \label{fig:product_image}}
\end{figure}

See \cref{fig:product_image} to see the store name, collection section, product name, product description and product image in the context of the Glovo application. For details of how this data was collected, we refer to \cref{sec:data_collection}.

\subsection{Data Collection}\label{sec:data_collection}
We collect data from all Glovo app partners. This data has been obtained and automatically saved in the app's databases during the six years that the app has been in operation. Most of the data included in the dataset is generated organically. That is, Glovo app does not enforce strict compliance measures for the names of products, descriptions and collection sections included in the catalog. This means that the partners that offer their menus through the app can freely generate this information as long as it is not offensive or breaks content rules.

The data featured in the dataset includes only those products present in the app that have an image associated with them. We decided to not include samples that contained no image information in order to have a fully multi-modal dataset. To maximise the usefulness of the dataset, we computed a hash of our samples (including an image hash) and deleted all identical samples. This reduced the dataset size from its original size of 7.5M samples to 2.8M. The reason behind the widespread presence of duplicated samples is the presence of large franchises in the markets where Glovo operates.

In this dataset we only include grocery partners that have food items in their catalog. This means that we exclude the majority of partners belonging to a new business vertical: q-commerce (``quick commerce"). These are mostly e-commerce partners with very large product selections and much less product variation across languages and countries. We intend to offer a curated version of this data in a future version of the dataset. It is worth mentioning that all data presented here has been publicly available at some point through the Glovo app. This means that it could potentially have been collected through crawling techniques by third-party actors. Whether this has happened or not, this dataset has not been made public through other sources.

In order to maximise the usefulness of our dataset, we decide to engage in minimal data postprocessing. This is limited to (i) reducing the maximum size of the product images, (ii) processing the store names in the dataset so that they can be more useful for ML practitioners. 

The images present in FooDI-ML are scaled so that the largest size of the image is always equal to or smaller than 512 pixels while maintaining the aspect ratio. No other transformation is implemented.

We also perform cleaning and processing of our store names. The original dataset had a large number of stores that were used by Glovo agents to backup store information after deactivation and to perform other tasks pertaining to the business' operations. This means that in this case the store name field does not add additional information (as it normally contains dummy information such as ``test" or ``backup"). It also means that hierarchical information obtained from these stores is not always useful (for example, if one wants to understand the relation between the menu and store name). To address this, we rename these store names with a generated string: ``AS\_N". This stands for ``Auxiliary Store number N" - where N is a positive integer.

\subsection{Summarised dataset statistics}
\subsubsection{Geography and Languages}
Our dataset includes samples belonging to up to 37 different countries in Europe, the Middle East, Africa and Latin America. Some languages such as Spanish (most of Latin America), English (in Nigeria, Ghana, Uganda), Russian (in several countries in Eastern Europe), Portuguese (Brazil, Portugal) are prevalent due to their broad geographical presence. Others like Ukrainian, Georgian and Italian are present due to the big market share that those countries represent in the app. 

We obtain statistics on language using the fastText classifier \cite{joulin2016bag} and filtering for those languages that we know that are not present in the country of operation. See \cite{app_2021} for details.

\begin{figure*}[ht]
  \centering
    \includegraphics[height=6cm]{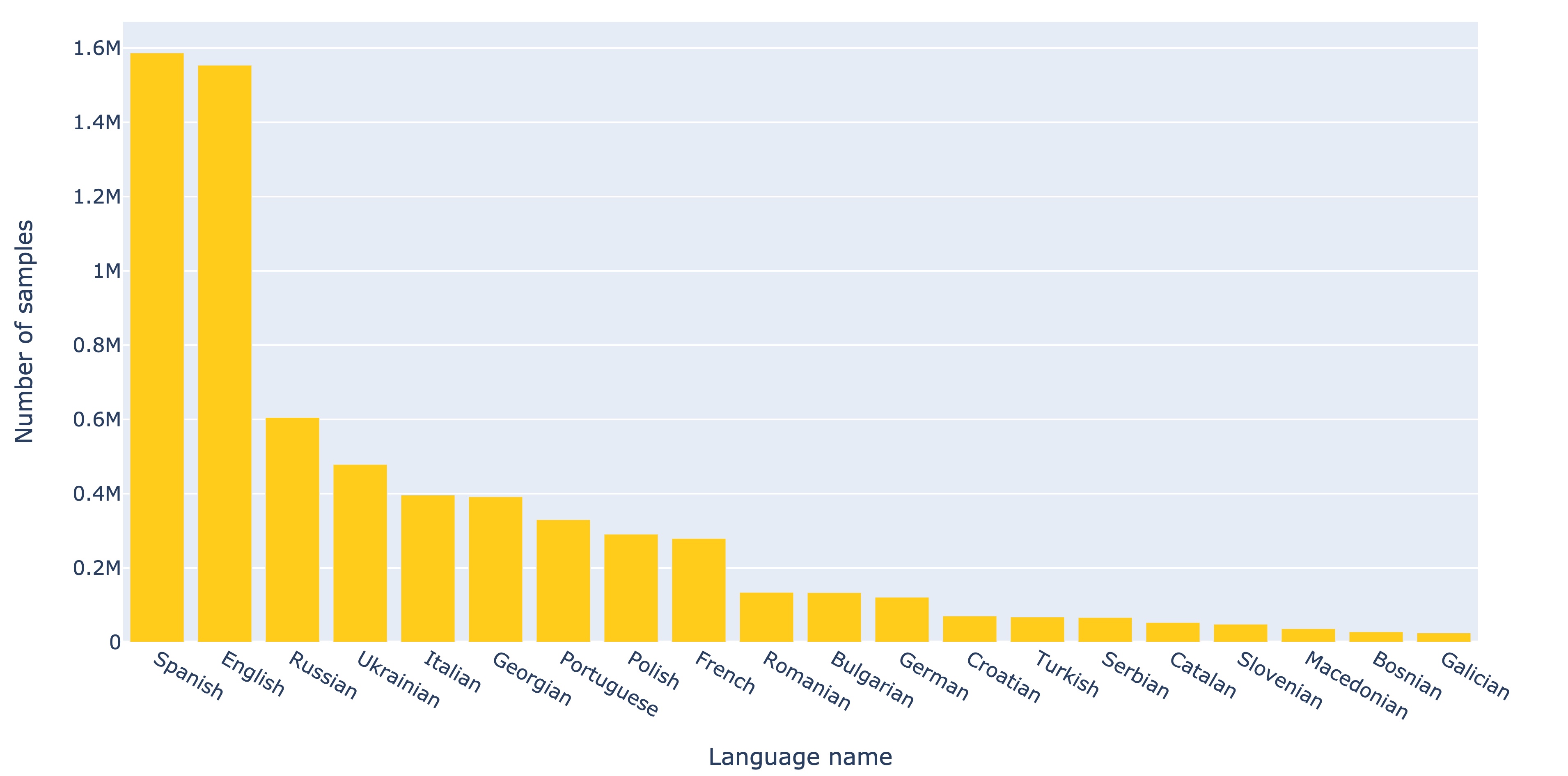}
  \caption{Presence of languages in our dataset in raw samples. Shown the top 20 languages. \label{fig:language_presence}}
\end{figure*}

\subsubsection{Language fields}
The Glovo app does not enforce the presence of product descriptions in the menus of the partners that use the app. This results in only $\sim$34\% of the samples (corresponding to $\sim$980k samples) including the product description field. The fact that some samples do not include product descriptions should not discourage researchers. In most cases, the product name and collection section are sufficient to fully describe the product, and concatenating these two fields provides a satisfactory natural language description of the product (see the analysis in \cite{app_2021}).

\subsubsection{Store name distribution}
Our dataset features samples from small restaurants and shops but also from large multinational food franchises. This means that the 2.8M samples are distributed amongst only 38K unique store names from which 36K (the overwhelming majority) feature less than 150 samples. If we disregard store names marked as Auxiliar Stores we see that 48\% of all samples belong to store names featuring less than 150 samples. That is, local restaurants and shops. The remaining 52\% belongs to larger chains or restaurants with very large, changing menus. This points towards a well-balanced dataset where the long tail of stores holds a significant percentage of the samples. 

\begin{figure*}[ht]
  \centering
    \includegraphics[height=6cm]{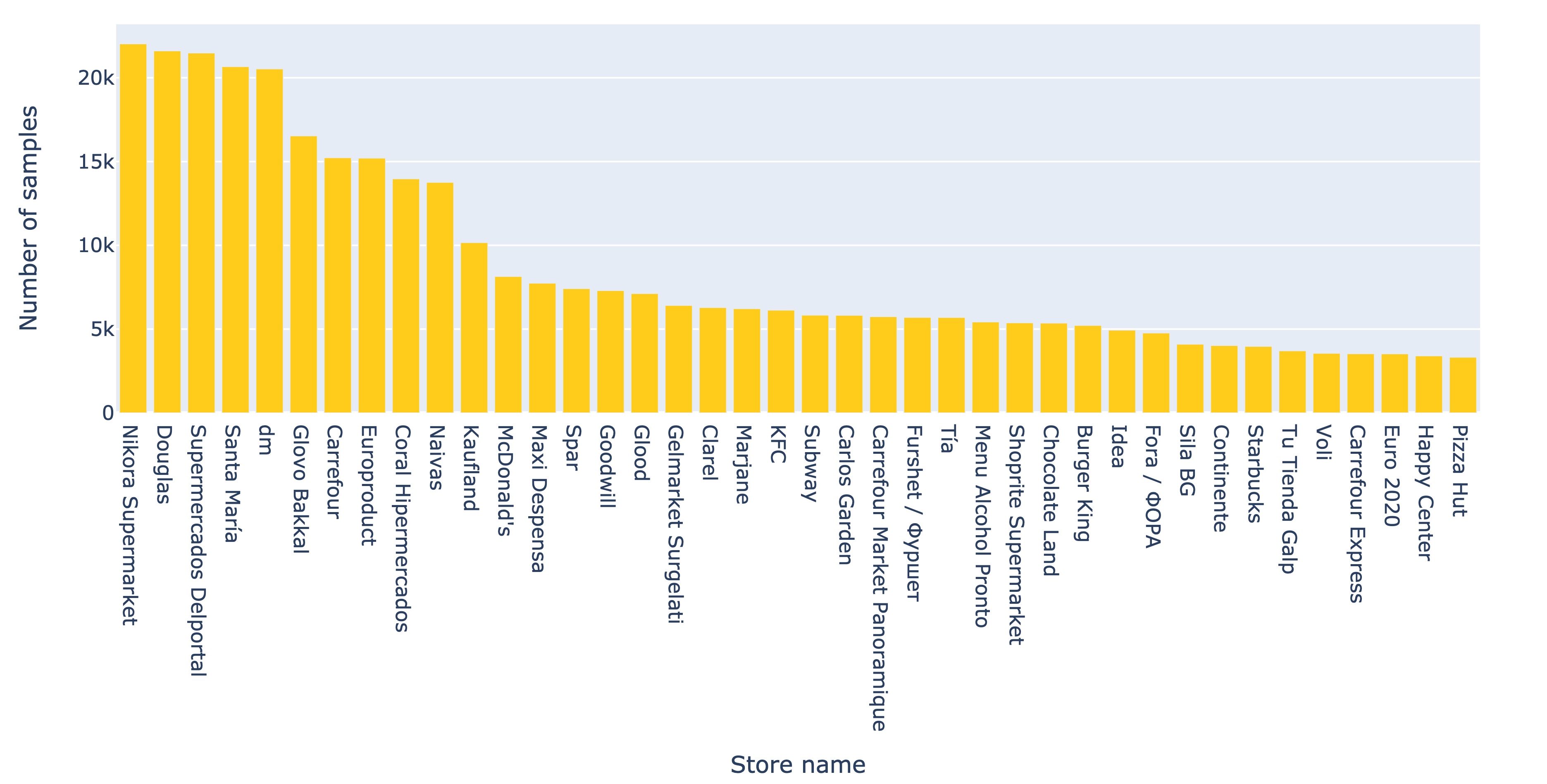}
  \caption{Store names with a larger number of samples, excluding auxiliary stores. Only shown the top 40 store names - representing around 9\% of the samples. Note how the dataset is well balanced between stores with a lot of samples and stores featuring a much more modest collection  \label{fig:stores_plot}.}
\end{figure*}

\cref{fig:stores_plot} shows the stores with more samples in descending order. As one would expect, grocery chains and large multinational brands top the list. This is due to the fact that grocery shops have much larger catalogs than restaurants, and that international brands may have the same product repeated many times over in different locations and different languages under the same store name.

\subsection{Notable examples and proposed tasks}
\subsubsection{Differential food samples}
\begin{figure*}[htb]
        \centering
        \begin{subfigure}[b]{0.35\textwidth}
            \centering
            \includegraphics[width=\textwidth]{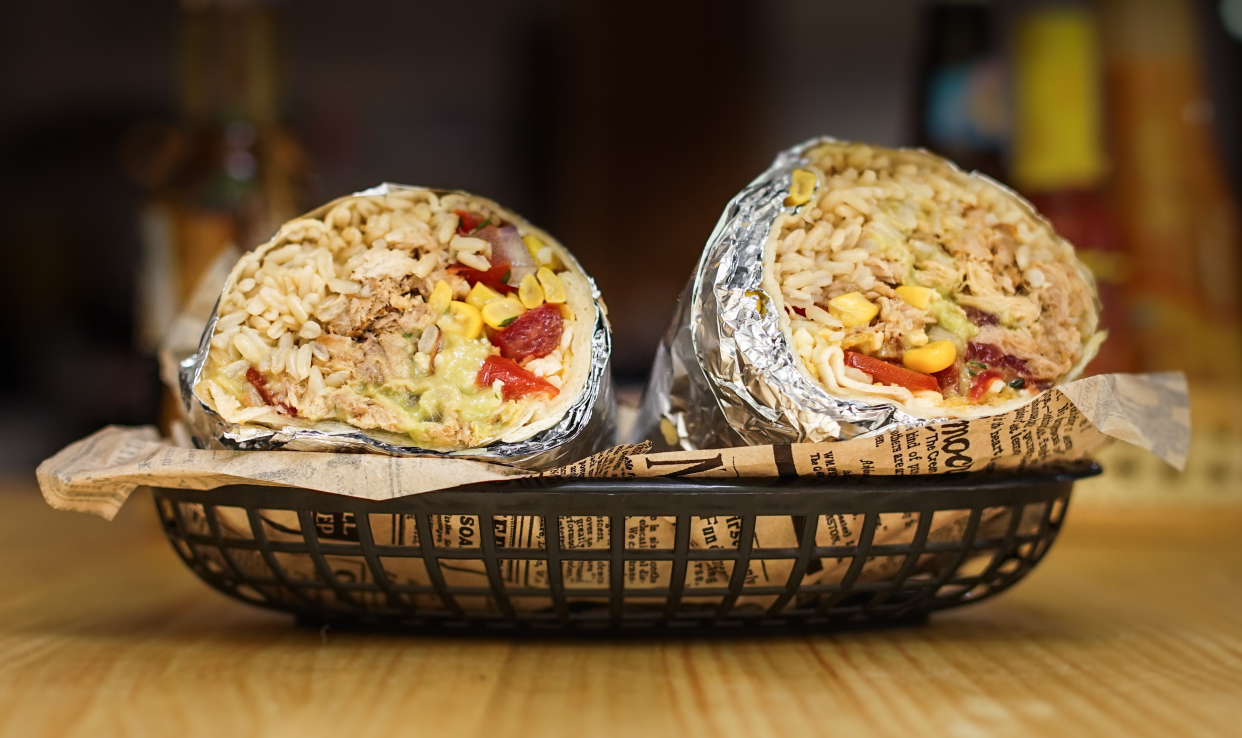}
            \caption[Network2]%
            {{ Burrito Carnitas. Product description: White or brown rice (your choice), carnitas (pork), guacamole, pico de gallo, grated cheese, corn and chips (minced) and jalapeños (optional).}} 
            \label{fig:bcarnitas}
        \end{subfigure}
        \begin{subfigure}[b]{0.33\textwidth}  
            \centering 
            \includegraphics[width=\textwidth]{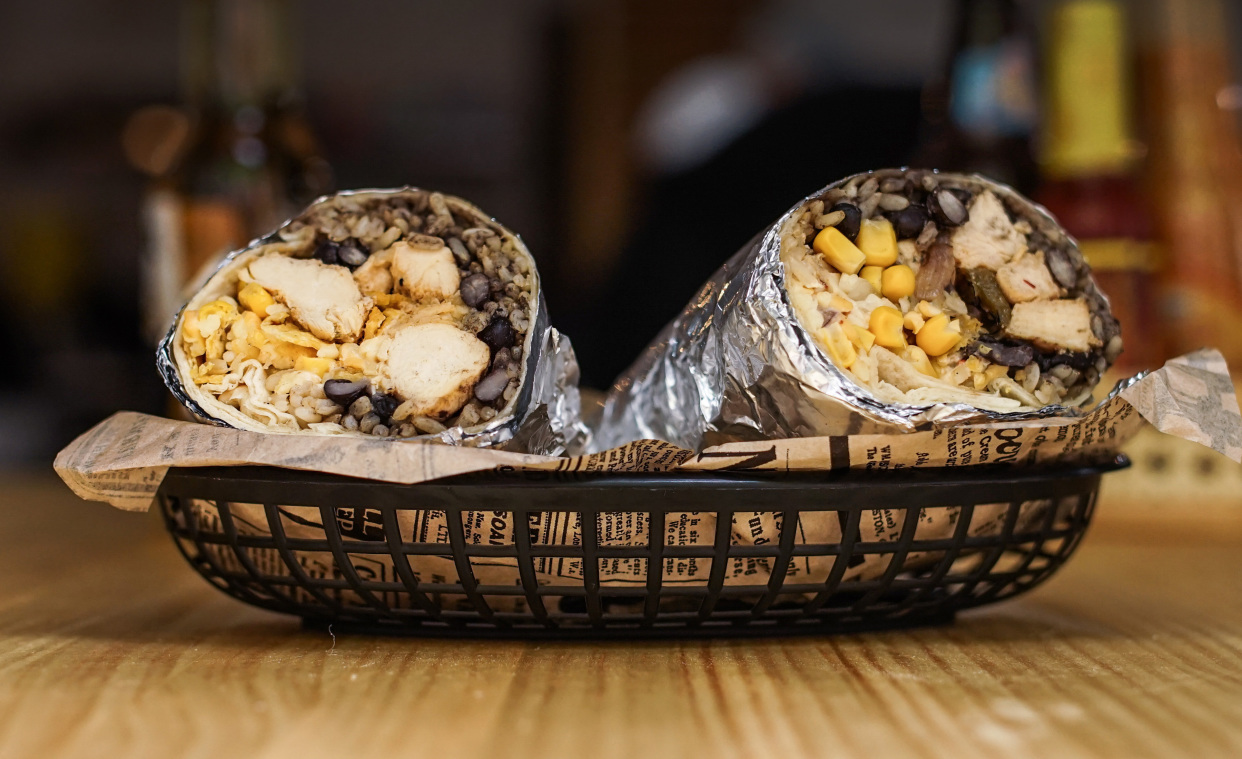}
            \caption[]%
            {{ Burrito Pollo. Product description: White or brown rice (your choice), beans, sauteed vegetables, chicken, chipotle sauce, grated cheese, corn and chips (minced) and jalapeños (optional).
            }}    
            \label{fig:bpollo}
        \end{subfigure}
        \vskip\baselineskip
        \begin{subfigure}[b]{0.35\textwidth}   
            \centering 
            \includegraphics[width=\textwidth]{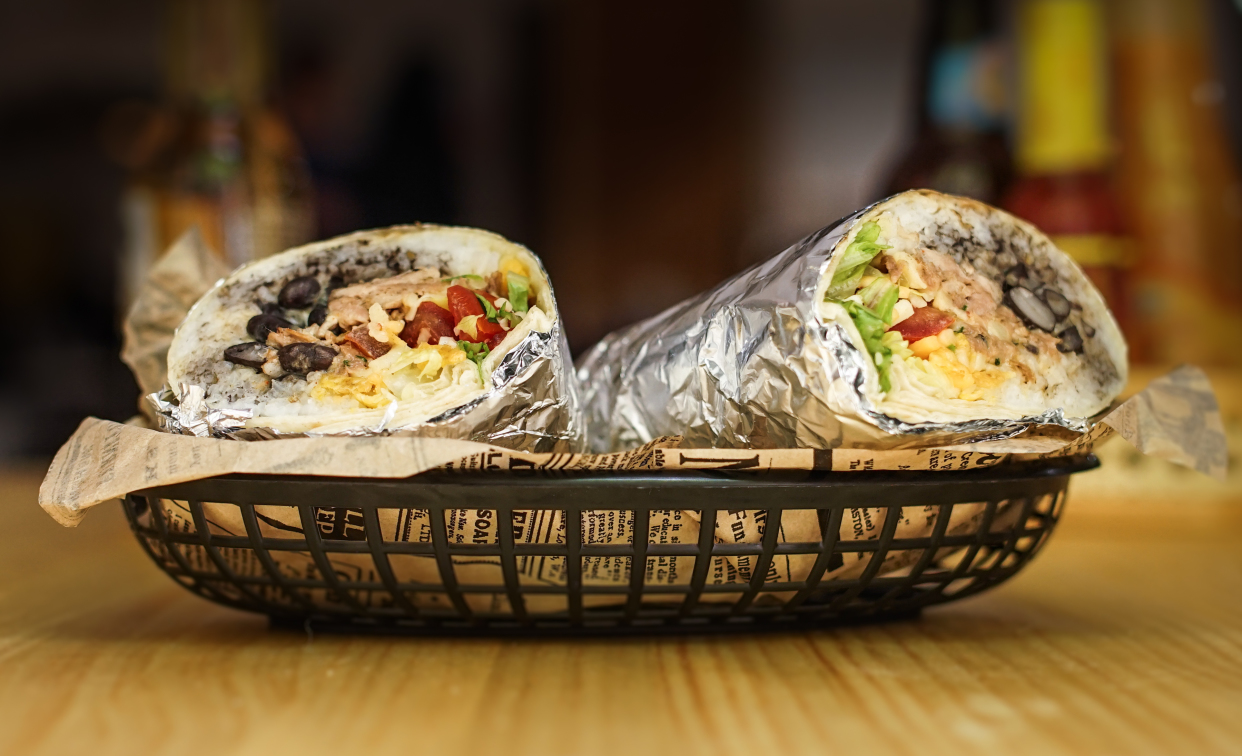}
            \caption[]%
            {{ Burrito Ternera BBQ. Product description: White or brown rice (your choice), beans, cheese sauce, BBQ beef, sour cream, pico de gallo, grated cheese, lettuce and chips (minced) and jalapeños (optional).
            }}    
            \label{fig:bternera}
        \end{subfigure}
        \begin{subfigure}[b]{0.35\textwidth}   
            \centering 
            \includegraphics[width=\textwidth]{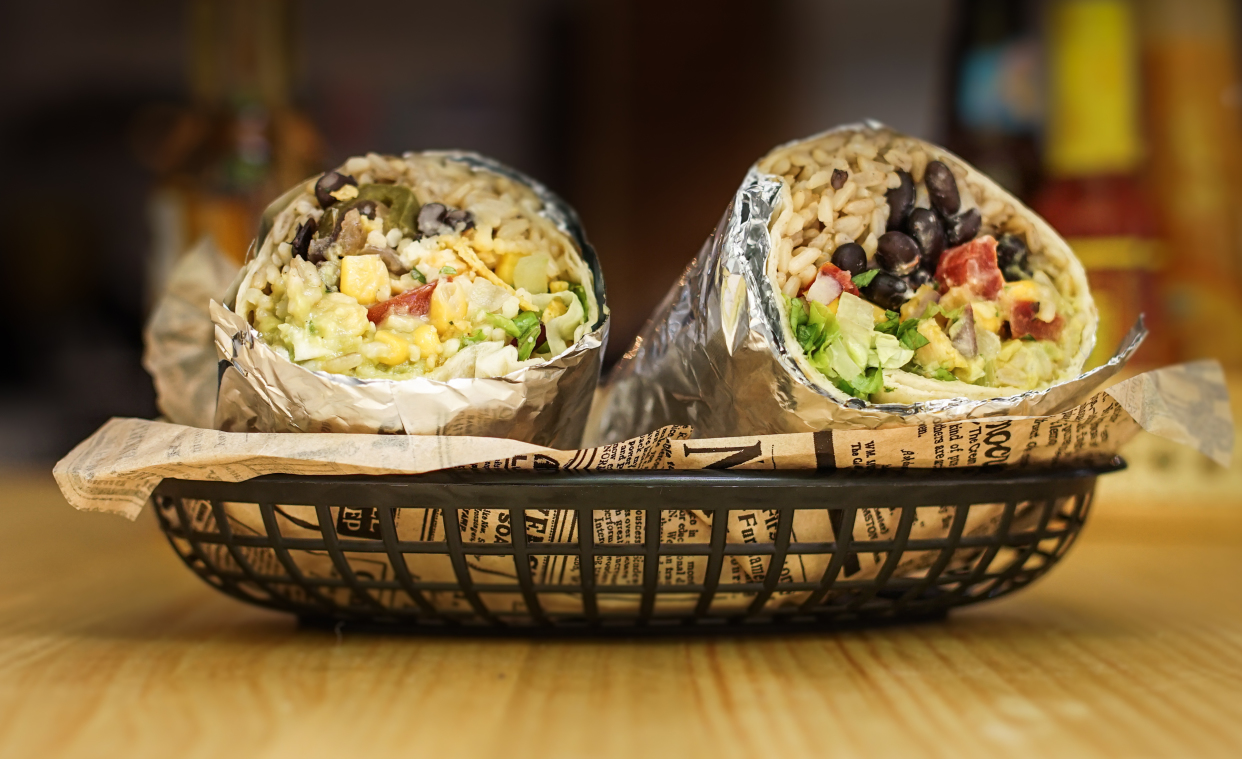}
            \caption[]%
            {{ Burrito Vegetal. Product description:
            White or brown rice (your choice), beans, sauteed vegetables, chipotle sauce, guacamole, pico de gallo, corn, lettuce and chips (minced) and jalapeño (optional)
            }}  
            \label{fig:bvegetal}
        \end{subfigure}
        \caption[Differential samples]
        { Figure showing what we call a \emph{differential food example}. That is, the same typology of plate (here, a burrito) with different ingredients. The photos are taken in a way ideal for saliency or segmentation approaches. Detailed descriptions are included, ideal for multi-modal learning (shown here translated from Spanish).} 
        \label{fig:burritos}
    \end{figure*}

We name ``differential food samples" the groups of samples that essentially contain the same entity but with different ingredients. This kind of samples coming from the same partner usually have similar image features such as illumination, background, object pose, etc. For example: in a burrito restaurant this can correspond to several pictures of the same burrito containing different combinations of ingredients - see \cref{fig:burritos}. This can be used to train saliency neural networks, ingredient generation algorithms, segmentation algorithms, etc.

Differential food samples are common in our dataset, as most pizzerias, burger stores, burrito stores,  salad places, etc. featuring descriptions include samples of this type. Although hard to precisely compute, we estimate that at least ~11\% of our samples that include a description belong to any of these food types and feature more than one sample per store (see \cite{app_2021} for the analysis). This produces tenths of thousands of examples, enough to train an ML algorithm.

\subsubsection{Text on image samples}
\begin{figure}[ht]
  \centering
    \includegraphics[angle=90,height=3cm]{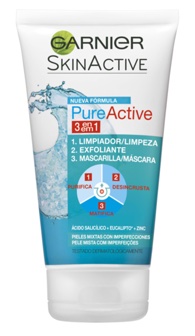}
  \caption{Image of a ``text on image sample" (rotated 90 degrees). The product name coincides partially with the image text. product name: Gel Limpiador Pure 3 En 1 Garnier 150 Ml  \label{fig:example_cleaning}.}
\end{figure}

Another interesting subset of samples present in our dataset are grocery samples, and samples from well-known fast-food companies. Many of such images are high-quality enough to be usable in optical character recognition tasks. Often, the text imprinted on the image also appears in the product description allowing for some interesting applications such as solving saliency tasks - see  \cref{fig:example_cleaning}. 

\subsubsection{Cross-language samples}
\begin{figure*}[htb]
        \centering
        \begin{subfigure}[b]{0.475\textwidth}
            \centering
            \includegraphics[width=0.5\textwidth]{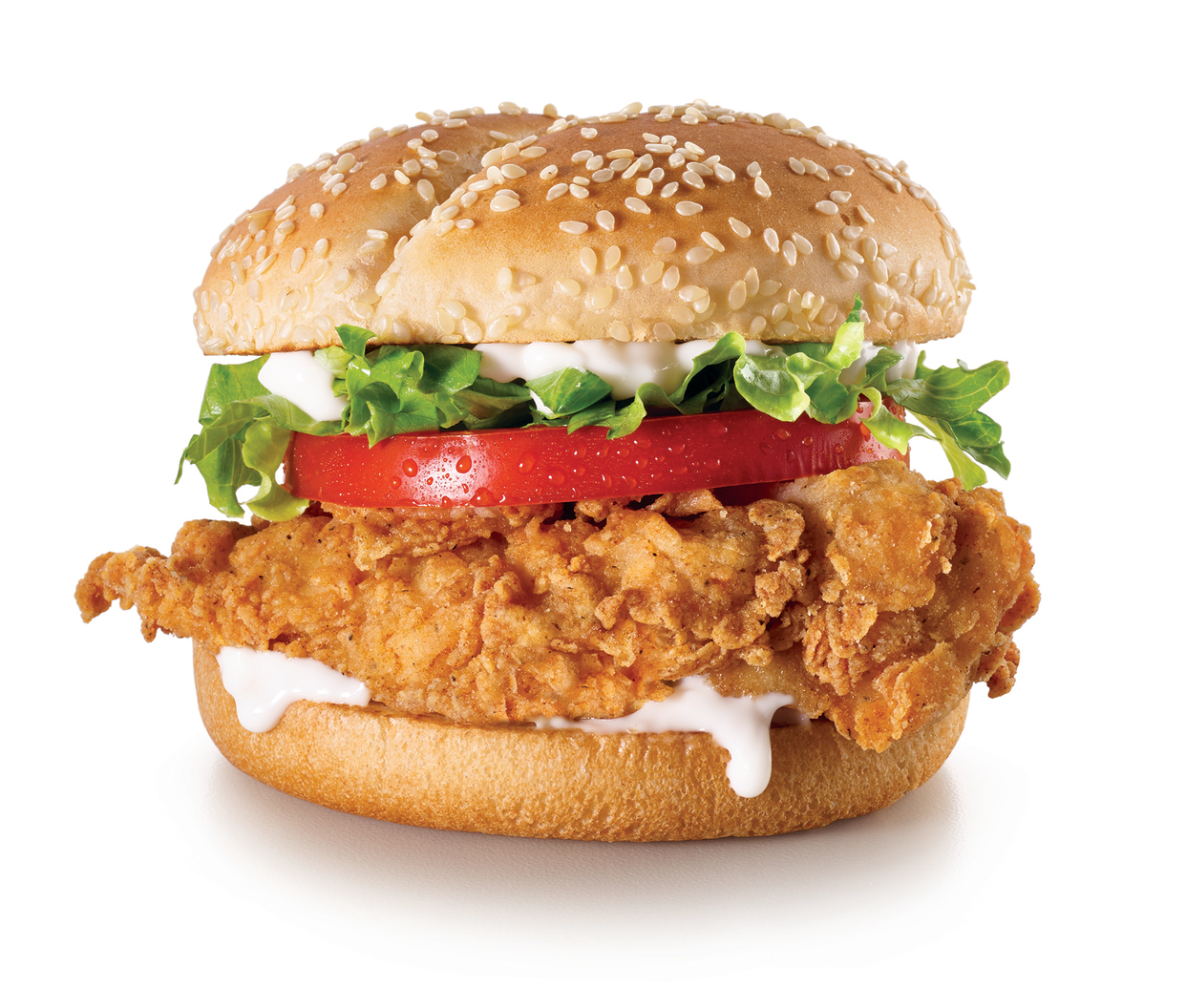}
            \caption[Network2]%
            {{ Colonel Burger. Product description: A tender and tasty marinated fillet cooked golden-brown, topped with fresh lettuce, a slice of tomato and Colonel dressing on a burger bun. Origin: Uganda.}} 
            \label{fig:uganda}
        \end{subfigure}
        \begin{subfigure}[b]{0.475\textwidth}  
            \centering 
            \includegraphics[width=0.5\textwidth]{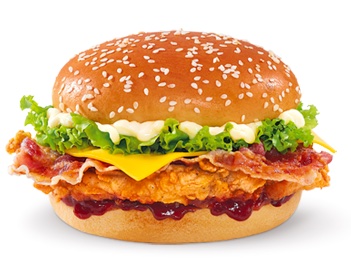}
            \caption[]%
            {{ Colonel's Burger. Product description: 1 chicken filet original recipe, BBQ sauce, bacon, cheddar, salad, mayonnaise. Origin: Italy. Translated from Italian.
            }}    
            \label{fig:italy}
        \end{subfigure}
        \vskip\baselineskip
        \begin{subfigure}[b]{0.475\textwidth}   
            \centering 
            \includegraphics[width=0.5\textwidth]{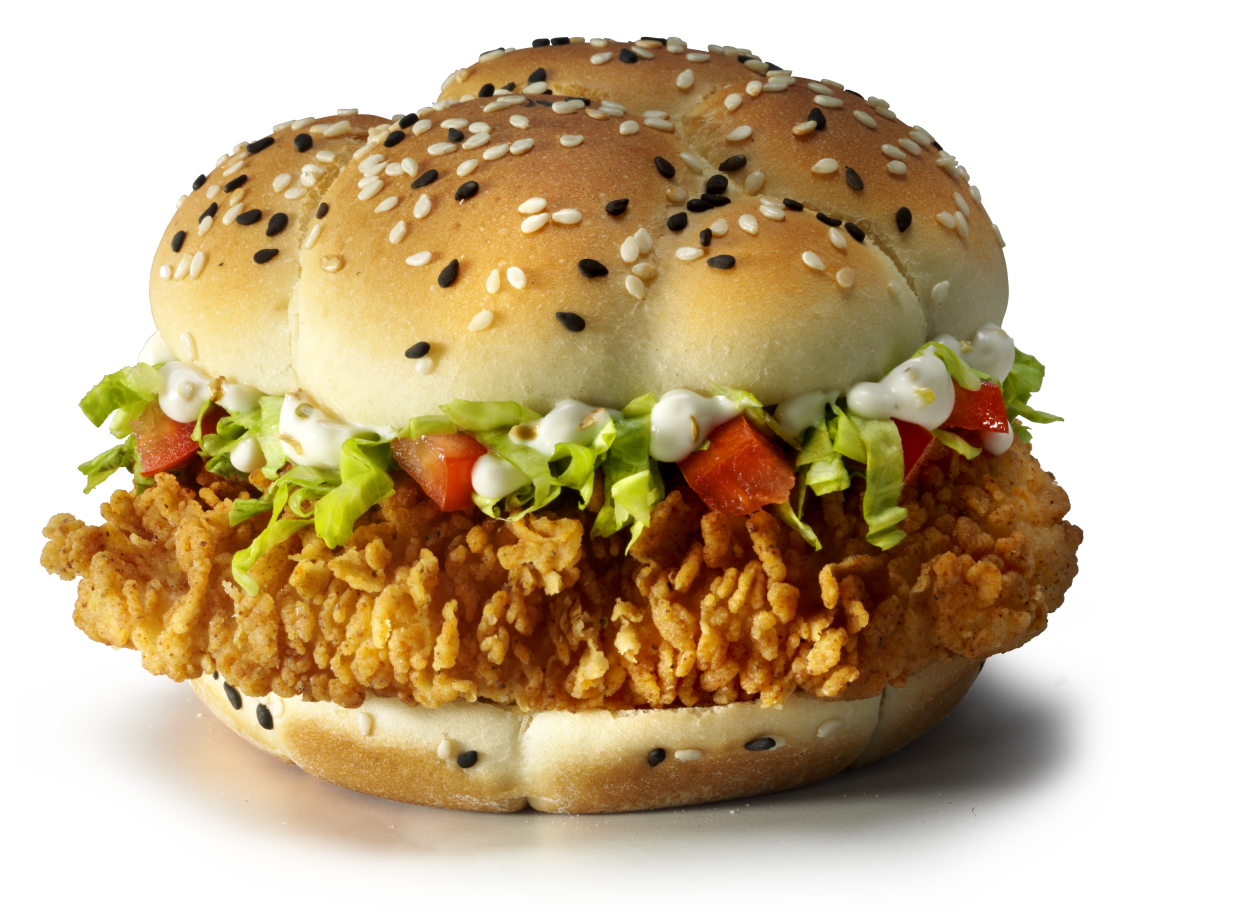}
            \caption[]%
            {{ Chef's Burger. Product description: Try a new unique burger from the chef! Delicate creamy sauce, juicy fillet in original breading, iceberg lettuce and tomatoes on a wheat bun with black and white sesame seeds. Origin: Kazakhstan. Translated from Russian.
            }}    
            \label{fig:kazak}
        \end{subfigure}
        \begin{subfigure}[b]{0.475\textwidth}   
            \centering 
            \includegraphics[width=0.43\textwidth]{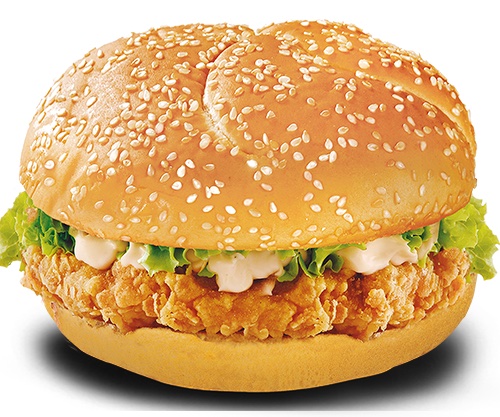}
            \caption[]%
            {{ Coronel burger. Sandwich with breaded chicken using a crispy recipe, Honduran sauce and lettuce. Origin: Honduras. Translated from Spanish.
            \\
            \\
            }}  
            \label{fig:honduras}
        \end{subfigure}
        \caption
        { Figure showing a cross-language sample. That is, a very similar sample (a burger from one of Glovo's partners), present in four different languages and locations. Note how the photography style, ingredients and description vary depending on location and language. } 
        \label{fig:multilingual_example}
    \end{figure*}

One of the main contributions of this dataset is the multiplicity of languages that it covers. Especially useful is the fact that we offer product descriptions of similar images in different languages. As an example, one can easily find similar burgers for one of our partners in Russian, Spanish, Italian and English with slightly different images and descriptions. See \cref{fig:multilingual_example} as an example. It is hard to calculate the number of occurrences for this type of sample. However, similar to differential food samples, a lower bound can be established from stores that are globally present - such as the big international chains shown in Fig. 4. Our estimation returns a lower bound of 4.5\% of samples (see \cite{app_2021}). This does not account for samples from different partners that still represent the same food type (for example, a pizza Margherita).

\subsection{Limitations and possible negative impact }\label{sec:limitations}
Despite its large sample size and language coverage, our dataset has some shortcomings.
\begin{itemize}
    \item Some languages are still underrepresented due to a ``lingua franca" effect. This happens both in developed and developing countries. For example in Spain where Catalan and Basque have a much smaller presence than their proportion of speakers. In Nigeria, Ghana and Uganda English is dominant while local languages such as Bantu and Hausa are not present in our dataset.  In summary, the dataset succeeds including languages that are dominant in their countries (like Ukrainian) but underrepresented in public datasets, but struggles to appropriately include minority languages. This can lead to the typical problems associated with the lack of training data: under-performance, bias and lack of generalisation.
    \item The issue of representation mentioned above is heavily influenced by market dynamics (which population groups are perceived to be more likely to use the app). This means that there is an additional danger of under-representing dialects perceived as representative of lower purchasing power populations.
    \item Glovo app partners are given a high degree of freedom when choosing their images and text representations. This causes some problems: (i) a high proportion of the samples do not include product descriptions. In most cases this is not a problem as the product name and collection section include enough information to obtain linguistic embeddings. However, in some cases - such as very well-known local dishes - this can be a problem. (ii) there is no standardisation on product images, which means that in a minority of samples the same image is used to represent different products. We have observed this, for example, in some small pizzerias where an image with several pizzas is used to represent all pizza choices. (iii) due to the very large amount of data and the freedom given to Glovo app partners and agents, it is possible that there are some low-quality tags and descriptions.
    \item Although the dataset is well balanced between small and big volume partners, it is also true that fast food dominates the delivery sector. This means that pizzas, burgers, fries, etc. are over-represented in the dataset compared to the real local diet of the countries where Glovo app operates. Similarly, dishes that are harder to deliver like cocktails, ice-cream cakes, steak, etc. are less present than easily deliverable food like sushi. This could cause the typical underperformance issue favouring fast food and deliverable food in front of other alternatives.
    \item Some global food types, such as pizza, burgers, sushi and salads are present in all languages shown in \cref{fig:language_presence}. However, food types with a local footprint such as -for example- ``coca de recapte" (a traditional Catalan savory pastry) are only present in local languages. In this case, Catalan and Spanish. This can hinder the performance of some cross-language tasks.
    \item All images included in our dataset are commercial images. At Glovo, stores are given the freedom to take and upload pictures independently which leads to large image variability. That said, it is true that photos included in a commercial application will differ from images of food taken in a real setting. Therefore, models trained with this dataset might struggle to generalise to all kinds of food pictures. Expanding the dataset with some other datasets - such as food datasets that rely on web crawling - might help to mitigate the issue.
\end{itemize}

\subsection{Data Split}\label{sec:proposed_split}
We propose a train/test/validation split of 70/15/15 stratified across countries and samples including product descriptions. We choose this stratification because the presence of a product description is a good indicator of a high-quality visio-linguistic sample, therefore ensuring that the evaluation and test metrics are representative and in line with the average quality of the samples. After this split, the train dataset contains 2M samples, and the test and validation splits contain 433k samples each. \cref{tab:table3} shows the full statistics for the proposed split. 

\begin{table}[htb]
  \vspace{1mm}
  \centering
  \resizebox{\columnwidth}{!}{%
  \begin{tabular}{llll}
    \toprule
    \cmidrule(r){1-2}
{Parameter} & {Train} & {Valid.} & {Test} \\
    \midrule
    Unique samples        & 2021210 & 433117 & 433117              \\
    Unique images     & 1234794 & 355500 &355965             \\
    Samp. w/ prod. nam. \& coll. sec.          & 2021210 & 433117 & 433117    \\
    Samp. w/ prod. description               & 481871 & 125193 & 125109     \\
        Unique store names (inc. Aux.)         & 37833 & 29647 &  29731  \\
        Unique coll. sec. & 88628 & 57376 & 57736 \\
        Unique prod. descriptions & 481871 & 125193 & 125109 \\
        Unique prod. names & 1174586 & 342826 & 342834 \\
    \bottomrule
  \end{tabular}}
    \caption{Statistics of the train test and validation splits.}
  \label{tab:table3}
\end{table}

\section{Benchmark tasks}
\label{sec:benchmark_task}
\subsection{Text image retrieval}\label{sec:benchmark_tir}

Text image retrieval is a sub-task of cross-modal retrieval that consists of retrieving an image given a query text or set of captions \cite{wang2019camp}. This retrieval is typically done by minimising a predefined distance between  two vectors: one representing the image, and one representing the text. Amongst other applications, text-image retrieval models are used to locate appropriate descriptions for a given image.

In this work we use two existing SotA approaches: CLIP \cite{radford2021learning}, and an adaptation of the dual-tower method used in the WIT dataset (henceforth, WIT) \cite{srinivasan2021wit}, to provide benchmark metrics for our dataset. CLIP and WIT are very similar in that both depend on fine-tuning previously independently trained encoders (such as ResNet \cite{he2015deep} or a Transformer encoder). The main difference between the two is that CLIP is trained over a very large private dataset (400M samples). In addition, CLIP is trained maximising a symmetric binary cross-entropy loss while WIT uses only the first component of the loss (corresponding to image to text retrieval). We use CLIP in a zero-shot manner as proposed by the authors. WIT's implementation is not publicly available, so we re-write it and offer it publicly. Note that in our implementation we use a transformer model instead of bag of words to reflect recent advances in sentence encoding. 

We use retrieval performance (R@N) as evaluation metric. Note that we count as positive examples any retrieved sample containing the correct image hash (t2i) or text string (i2t). Both WIT and CLIP perform much better than random. Scores significantly vary depending on the model: surprisingly, CLIP (zero-shot) performs much better than the WIT approach (trained on our dataset). This serves as an additional experimental proof of the excellent generalization capabilities of the CLIP network. 

\begin{figure}[htb]
        \centering
        \begin{subfigure}[b]{0.475\textwidth}
            \centering
            \includegraphics[width=1\textwidth]{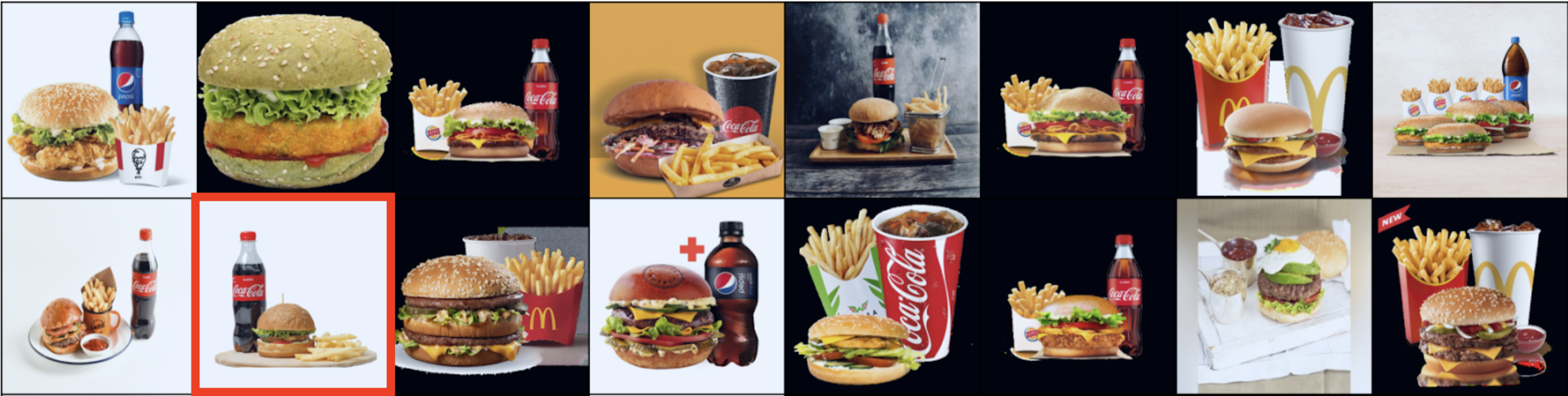}
            \label{fig:i2t_img}
                    \caption
        {Example of performance on the image to text retrieval task (illustrated with associated images). Image query marked in red. Here, the algorithm struggles differentiating amongst different burger descriptions. The appropriate text is ``beef hamburger + french fries + coca-cola (0.5 l.) pan america" (translated from Russian). The correct sample is at N=10.} 
        \end{subfigure}
        \begin{subfigure}[b]{0.475\textwidth}  
            \centering 
            \includegraphics[width=1\textwidth]{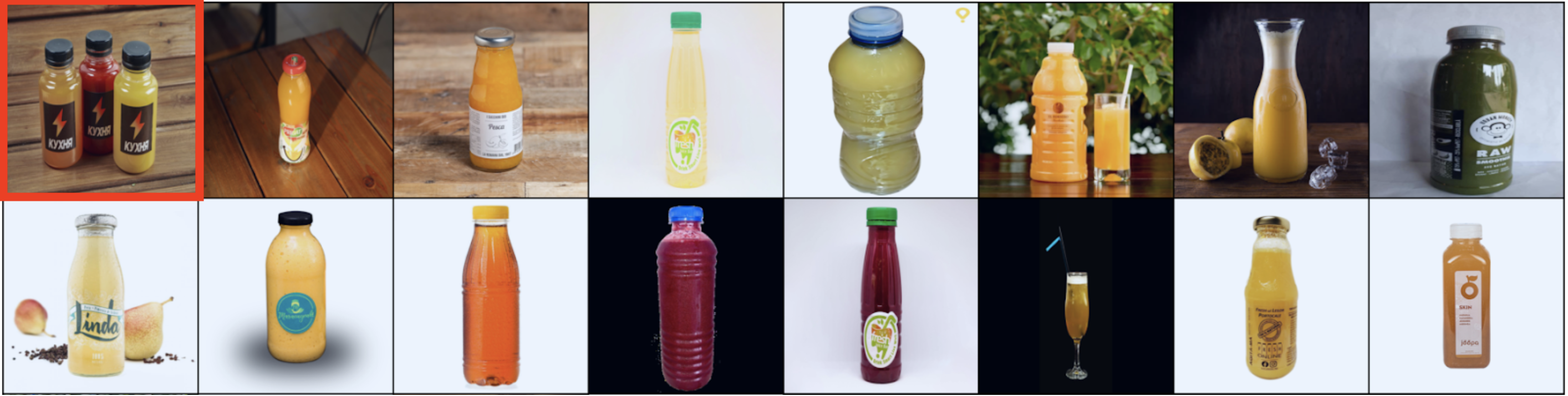}
                    \caption
        { Example of performance on the text to image retrieval task (illustrated with retrieved images). Query text (from Russian): ``tomato juice (250ml)". Here, the correct sample is at N=1 (marked in red). } 
            \label{fig:t2i_img}
        \end{subfigure}
       
        \caption
        {Examples of the performance of WIT on the text-image retrieval task. (a) illustrates i2t and (b) illustrates t2i. Note how the strong similarlity between images means that this is a hard task to solve for the trained network.   } 
        \label{fig:samples_retrieval}
    \end{figure}

The performance of the WIT and CLIP method on our dataset is significantly better than the metrics reported for similar datasets (see  \cite{srinivasan2021wit} for example) . However, it is much worse than the performances obtained for smaller datasets such as  flickr30K and COCO. This is to be expected as it is harder to retrieve the right sample the larger the dataset is. To illustrate the difficulty of the task, we include the result of several image to text and text to image queries (see \cref{fig:samples_retrieval}). It is clear from the figure that the algorithm -in this case, WIT- has learnt correctly. However, it is still very hard to retrieve the appropriate sample amongst a validation set of over 280k examples.


\begin{table}[ht]
  \centering
    \resizebox{\columnwidth}{!}{
  \begin{tabular}{lllllll}
    \toprule
    \cmidrule(r){1-2}
{Method} & {A-R@1} & {A-R@5} & {A-R@10} &{Re-R@1} & {Re-R@5} & {Re-R@10}  \\
    \midrule
                WIT        & 0.25           & 0.62           & 0.96 & 0.22 & 0.6 & 0.91            \\
                CLIP & 1.58           & 3.73           &  5.12 & 1.63 & 3.7 & 4.97\\
    \bottomrule
  \end{tabular}}
    \caption{Performance of the WIT and CLIP methods. ``A" stands for annotation (i2t) and ``Re" for retrieval (t2i) scores.}
  \label{tab:table4}
\end{table}

\subsection{Conditional image generation}\label{sec:benchmark_gan}
In addition to text image retrieval, we also train a Generative Adversarial Network (GAN) to conditionally generate pizza images. We choose this task amongst many other options to showcase how easy is it to (i) modify the dataset annotations into categorical variables, (ii) use the rich image dataset to train generative models.

\begin{figure}[htb]
        \centering
        \includegraphics[width=0.5\textwidth]{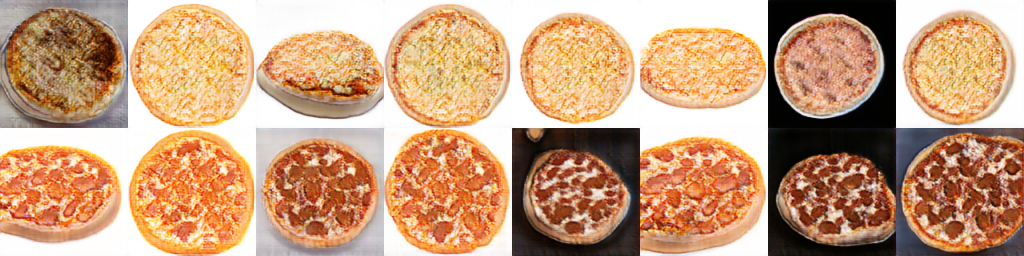}
        \caption
        { Automatically generated images for four cheese pizza (top) and pepperoni pizza (bottom).} 
        \label{fig:pepperoni_samples}
    \end{figure}

To train our GAN we use a Self-Attention GAN \cite{zhang2019self}, a well-established method allows for easy conditioning of generated images. Upon training, we obtain an inception score of 3.58 and realistic images of several common pizza types (here shown, pepperoni and four cheese - see \cref{fig:pepperoni_samples}).

\section{Conclusion and future work}
\label{sec:conclusion}
We present a large multilingual visio-linguistic dataset covering 33 languages, 37 countries, over 2.8M images and 9.5M text samples: FooDI-ML. The dataset is obtained from the partners of a large food and grocery delivery company: Glovo. FooDI-ML is a domain-specific dataset covering mostly the case of food, drinks and grocery products. The data is collected over six years of Glovo's operations.

This dataset opens the door for several applications, insofar unavailable for the broader research community due to the lack of public datasets. For instance, multilingual image-based search engines based on food and drink examples, refinement of existing pretrained models for the food and drink industry, and improvement of food image embeddings. The only comparable dataset in this domain is the recipe1M+ dataset, limited to the English language.


In addition to describing the dataset, we include an overview of notable samples. We also provide a train/test/validation split and several benchmarks over two different tasks. As future work, we plan to release the V2 of this dataset, including many more grocery and marketplace products. We estimate that the V2 of the dataset will at least double the size of the dataset that we present here. We also plan to release some specialized tasks for the food and drinks domain.

{\small
\bibliographystyle{ieee_fullname}
\bibliography{PaperForReview}
}

\end{document}